\documentclass{article}

\usepackage{microtype}
\usepackage{graphicx}
\usepackage{subcaption}
\usepackage{booktabs} 

\usepackage{hyperref}
\usepackage[table,xcdraw]{xcolor}
\usepackage{xcolor}
\usepackage{graphicx}
\usepackage{amsmath}
\usepackage{amssymb}
\usepackage{pifont}
\usepackage{booktabs}
\usepackage{multirow}
\usepackage{makecell}
\usepackage{overpic}
\usepackage{algorithm}
\usepackage{enumerate}
\usepackage{algpseudocode}

\usepackage{color}

\definecolor{turquoise}{rgb}{0.6,0.4,0}
\definecolor{orange}{rgb}{0.8, 0.6, 0.2}
\definecolor{brown}{rgb}{0.5, 0.16, 0.16}
\definecolor{black}{rgb}{0,0,0}
\definecolor{blue}{rgb}{0,0,1}
\definecolor{teal}{rgb}{0.0, 0.4, 0.4}
\definecolor{purple}{rgb}{0.65,0,0.65}
\definecolor{green}{RGB}{0,1,0}
\definecolor{yellow}{RGB}{238, 155, 0}
\definecolor{red}{rgb}{0.9, 0, 0}
\definecolor{gray}{rgb}{0.94, 0.94, 0.94}

\newcommand{\delete}[1]{}

\newcommand{\myparagraph}[1]{\vspace{1mm}\noindent\textbf{#1}~}

\usepackage[accepted]{icml2026}

\usepackage{amsmath}
\usepackage{amssymb}
\usepackage{mathtools}
\usepackage{amsthm}

\usepackage[capitalize,noabbrev]{cleveref}

\theoremstyle{plain}

\theoremstyle{definition}

\theoremstyle{remark}

\usepackage[textsize=tiny]{todonotes}

\usepackage{multicol}

\newbool{showComments}

\usepackage{xcolor}
\usepackage[normalem]{ulem} 

\ifbool{showComments}{%

\newcommand{\jxdel}[1]{\sethlcolor{pink}\hl{#1}}

\newcommand{\del}[1]{\textcolor{red}{\sout{#1}}}
}{

\newcommand{\jxdel}[1]{}

\newcommand{\del}[1]{}
}

\icmltitlerunning{TextMesh4D: Zero-shot Text-to-4D Mesh Generation}

\begin{document}

\twocolumn[
  \icmltitle{TextMesh4D: Zero-shot Text-to-4D Mesh Generation}



  \icmlsetsymbol{equal}{*}

  \begin{icmlauthorlist}
    \icmlauthor{Sisi Dai}{yyy}
    \icmlauthor{Xinxin Su}{yyy}
    \icmlauthor{Kai Xu}{xxx,zzz}
  \end{icmlauthorlist}

  \icmlaffiliation{yyy}{National University of Defense Technology}
  \icmlaffiliation{xxx}{Institute of AI for Industries, Chinese Academy of Sciences, Nanjing, 211135, China}
  \icmlaffiliation{zzz}{Jiangsu Key Laboratory of AI for Industries
Institute of AI for Industries, Chinese Academy of Sciences}

  \icmlcorrespondingauthor{Kai Xu}{kevin.kai.xu@gmail.com}

  \icmlkeywords{Machine Learning, ICML}

  \vskip 0.3in

\begin{center}
    {\includegraphics[width=0.82\linewidth]{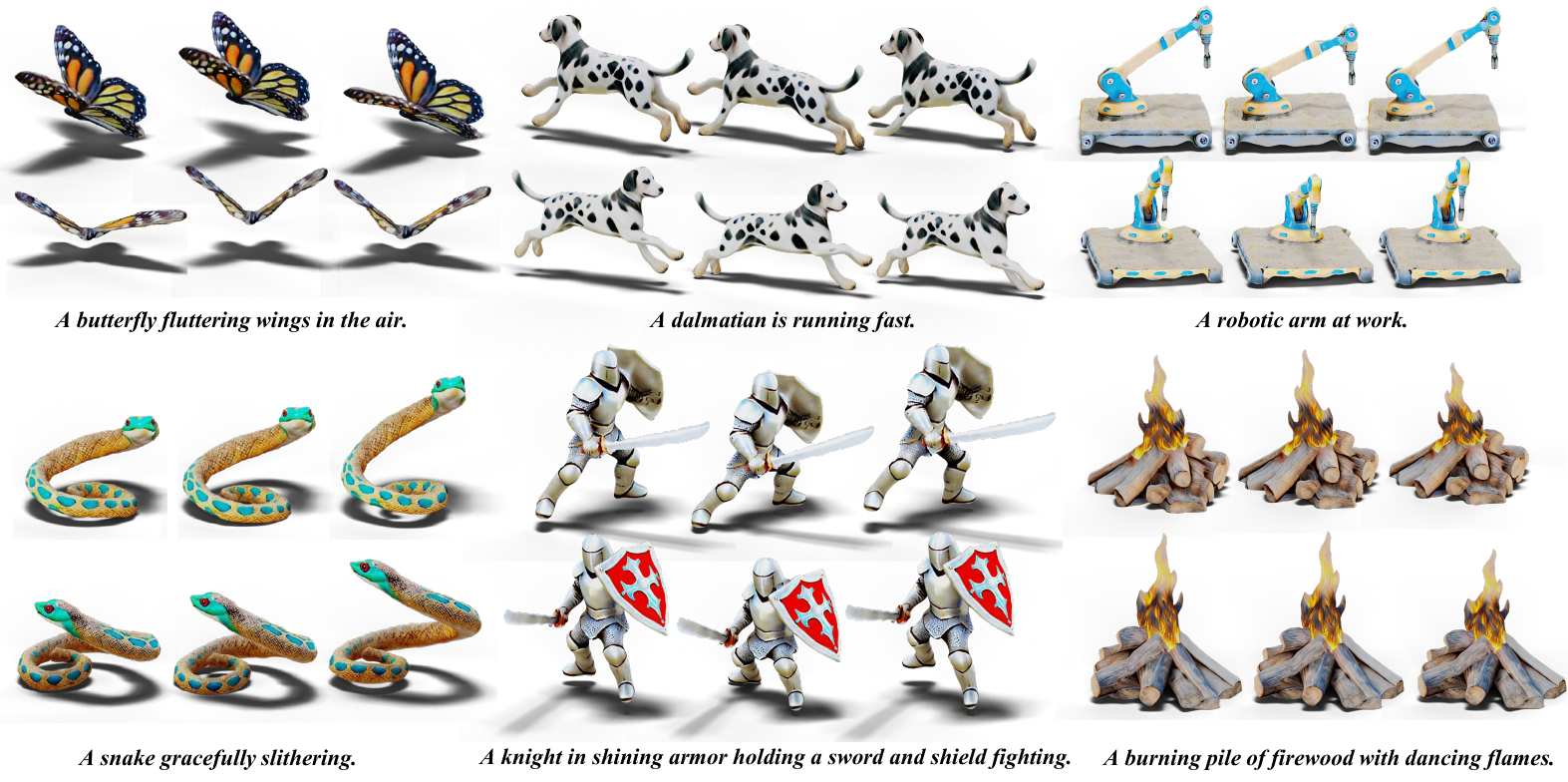}}
    \captionof{figure}{TextMesh4D generates diverse 4D meshes from a single text prompt.}
    \label{fig:diverse}
\end{center}
  
]



\printAffiliationsAndNotice{}  

\newcommand{\renderer}{g}
\newcommand{\modelparams}{\psi}
\newcommand{\renderedimage}{x}
\newcommand{\camerapose}{\zeta}
\newcommand{\textprompt}{T}
\newcommand{\textembedding}{y}
\newcommand{\timestep}{t}
\newcommand{\noise}{\epsilon}
\newcommand{\noisepredictnet}{\phi} 
\newcommand{\cameraextrinsics}{\mathbf{c}} 
\newcommand{\bodypose}{\theta}
\newcommand{\bodyshape}{\beta}
\newcommand{\spatialpoint}{p}
\newcommand{\allpoints}{P}
\newcommand{\crossentropy}{CE}
\newcommand{\hyperparameter}{\eta}
\newcommand{\distancetoanchor}{d}

\begin{abstract}

Large-scale, high-quality dynamic 3D (4D) assets are essential for learning physically grounded representations, but remain costly to capture and annotate at scale. This limits the viability of supervised 4D learning and motivates zero-shot text-to-4D generation leveraging pretrained diffusion priors. To model complex dynamics, prior methods typically adopt implicit 3D representations (e.g., NeRFs or 3DGS) for their deformation capacity. However, their implicit nature provides limited control over surface topology, which hinders high-fidelity geometry and makes temporally coherent surface reconstruction challenging. To address these limitations, we explore zero-shot text-to-4D mesh generation. However, a structural mismatch arises when combining diffusion-based guidance with topology-constrained meshes: the guidance is noisy and spatially inconsistent, while meshes impose severe topological constraints, making direct vertex-level deformation unstable.
In this paper, we introduce \emph{TextMesh4D}, the first zero-shot framework for text-to-4D that directly generates dynamic meshes by addressing the above challenge at two complementary levels. \textbf{Geometrically}, we shift deformation modeling from vertices to faces via a \emph{Jacobian Deformation Field (JDF)}, enabling topology-aware surface reconstruction through an integrability-enforcing integration formulation. \textbf{Semantically}, we propose a \emph{Local-Global Semantic Regularizer (LGSR)} that preserves identity over time by jointly constraining local deformation plausibility and global shape consistency. Extensive experiments demonstrate state-of-the-art temporal consistency, structural fidelity, and visual quality, while remaining efficient on a single 24GB GPU.

\end{abstract}
\section{Introduction}

Artificial General Intelligence (AGI) ultimately demands grounded physical understanding, where 3D objects evolve coherently over time. Learning such understanding relies on large-scale and semantically diverse dynamic 3D (i.e., 4D) representations, yet existing high-quality 4D data remains severely limited in both scale and diversity. This scarcity fundamentally impedes conventional supervised learning, motivating scalable 4D generation as a practical alternative to data acquisition. As a result, recent research has increasingly shifted toward zero-shot text-to-4D generation that leverages pretrained image and video diffusion priors.

Following this emerging trend, recent advancements typically follow a two-stage pipeline: (i) static 3D synthesis followed by (ii) motion generation guided by diffusion priors. While the first stage can now produce high-fidelity static 3D geometry, the subsequent motion generation remains a key bottleneck, struggling to reconcile \textbf{\emph{motion flexibility}} with \textbf{\emph{geometric fidelity}}. To model complex dynamics, prior work commonly resorts to implicit volumetric representations (e.g., NeRFs~\cite{nerf} or 3D Gaussian Splatting~\cite{3dgs}) for their deformation capacity. However, this design choice introduces notable practical limitations. 
First, without explicit topological constraints, implicit fields can overfit the stochastic, high-frequency signals induced by diffusion-based supervision, leading to spurious variations in density and appearance. Second, relying primarily on multi-view visual consistency provides limited structural regularization. As a result, it is difficult to reliably extract temporally coherent, surface-based 4D assets, limiting compatibility with standard graphics pipelines.

To bridge this gap and enable direct graphics compatibility, a natural question arises: \emph{is there a topology-aware solution that leverages explicit surface topology (e.g., meshes) to better reconcile motion flexibility with geometric fidelity?} Despite offering explicit topology and broad support in graphics, mesh-based representations remain largely underexplored for zero-shot text-to-4D generation. To examine their feasibility, we conduct an extensive empirical study of diffusion-guided vertex-level deformation~\cite{clipmesh,textmesh} driven by video diffusion priors, spanning a wide range of parameterizations and regularizers. Across these settings, optimization consistently fails to recover coherent surfaces, exhibiting artifacts such as self-intersections and geometric collapse, especially under significant motion. We attribute these failures to the noisy and spatially inconsistent gradients from diffusion priors: neighboring vertices often receive conflicting supervision, which destabilizes mesh connectivity. This reveals a fundamental mismatch between naive vertex-based deformation and stochastic generative supervision.

We further analyze this mismatch as a structural conflict between diffusion supervision and vertex-based meshes. Granting vertices sufficient freedom amplifies artifacts from inconsistent gradients, whereas enforcing smoothness to stabilize optimization suppresses expressive motion. To resolve this dilemma, our key insight is to shift deformation modeling \textbf{\emph{from the spatial domain of vertices to the differential domain of faces}}, thereby changing how noisy supervision is absorbed. In contrast to vertices, each face is allowed to rotate and stretch locally to absorb noise, whereby global coherence is restored by formulating surface reconstruction as a constrained integration problem. This formulation exploits mesh topology to suppress spatially incoherent signals and propagate consistent deformation trends, thereby mitigating the impact of noisy priors.

Given this insight, we propose \emph{TextMesh4D}, the first zero-shot text-to-4D framework that generates \emph{dynamic meshes} under diffusion guidance. TextMesh4D addresses the mismatch between stochastic diffusion supervision and topology-constrained meshes from two complementary aspects. \textbf{Geometrically}, instead of optimizing vertices, we model deformation in the face domain via a \textbf{\emph{Jacobian Deformation Field (JDF)}} that assigns a learnable per-face Jacobian to capture local rotation and stretch; a globally consistent surface is then recovered by Poisson integration, which enforces integrability and leverages mesh topology to suppress spatially incoherent gradients. \textbf{Semantically}, we introduce a \textbf{\emph{Local-Global Semantic Regularizer (LGSR)}} to stabilize identity over time: a local ARAP-style term discourages non-physical, fluid-like distortions, while a global canonical-shape anchoring term prevents long-horizon drift. Together, JDF and LGSR enable temporally coherent, graphics-ready 4D meshes while retaining expressive non-rigid motion under zero-shot generative supervision.

 Our contributions are summarized as follows:

\begin{itemize}
    \item We propose \textbf{TextMesh4D}, the first zero-shot text-to-4D generation framework that directly outputs \emph{dynamic meshes}.
    \item We introduce the \textbf{Jacobian Deformation Field (JDF)}, a face-domain deformation parameterization with Poisson-based recovery that enforces integrability and yields topology-consistent 4D surfaces under diffusion supervision.
    \item We propose the \textbf{Local-Global Semantic Regularizer (LGSR)} to jointly preserve local physical plausibility and global long-term identity, mitigating semantic drift in diffusion-guided optimization.
    \item Extensive experiments demonstrate state-of-the-art temporal consistency, structural fidelity, and visual quality, with efficient training on a single 24GB GPU.
\end{itemize}

\paragraph{Conflict of Interest Disclosure.} The authors declare no conflict of interest.
\section{Related Work}

Our framework leverages diffusion priors from text-to-image/video generation and the two-stage text-to-3D/4D design, while targeting explicit 4D mesh assets rather than implicit 4D fields. We next review related work in (i) text-to-image/video generation, (ii) text-to-3D generation, and (iii) text-to-4D generation.

\paragraph{Text-to-Image/Video Generation.}
In recent years, diffusion models have achieved significant advancements in image and video generation, including text-to-image (T2I) models~\cite{dall-e, stable-diffusion, imagen}, as well as text-to-video (T2V) models~\cite{ModelScopeT2V, videocrafter1, zeroscope}. These models are trained on large-scale open-domain datasets, typically including LAION-5B~\cite{laion-5b}, WebVid-10M~\cite{WebVid-10M}, and HD-VG-130M~\cite{HD-VG-130M}. Recent advancements in text-image-to-video generation (TI2V) have incorporated image-based semantic conditions into T2V models~\cite{seer, videocomposer, I2VGen-XL}. The latest model DynamiCrafter~\cite{dynamicrafter} employs a learnable image encoding network and dual cross-attention layers to effectively integrate text and image information, achieving impressive open-domain TI2V generation. Our work distills the generative power of video diffusion models for motion generation, with the belief that our method will evolve accordingly with the continued advancement of video generation technology.

\paragraph{Text-to-3D Generation.}
Early methods \cite{chen2019text2shape,jahan2021semantics,liu2022towards} for text-to-3D generation require paired data of 3D data and corresponding textual descriptions, which limits their generality to unseen object categories. Benefiting from large pre-trained text-to-image models and differentiable rendering techniques, breakthroughs~\cite{dreamfields, pureclipnerf, clipmesh, avatarclip} in text-to-3D content generation have been achieved. Recently, the technique SDS (Score Distillation Sampling) has been introduced in the pioneering work DreamFusion \cite{dreamfusion}, enabling 3D generation by distilling guidance from pre-trained T2I diffusion models. There are a lot of follow-up works to improve DreamFusion. Some focus on 3D representation~\cite{magic3d, fantasia3d}: Magic3D \cite{magic3d} proposes a coarse-to-fine pipeline to generate the fine-grained mesh; TextMesh \cite{textmesh} extends the geometry representation from NeRF to an SDF framework, enhancing detailed mesh extraction and photorealistic rendering; DreamGaussian~\cite {dreamgaussian} proposes to adopt 3D Gaussian Splatting to increase efficiency. Some works focus on improving SDS: SJC~\cite{sjc} proposes a variant of SDS, while VSD are proposed in ProlificDreamer~\cite{prolificdreamer}; DreamTime~\cite{dreamtime} improves the generation quality by modifying the timestep sampling strategy. Others focus on inducing 3D priors into the guidance source, which effectively alleviates the Janus problem. Additional 3D priors are introduced in shape~\cite{latentnerf, dream3d, dreamavatar, dreamwaltz, interfusion, compositional}, providing geometric initial values for optimizing NeRF. MVDream~\cite{mvdream} proposes to fine-tune the diffusion model to generate multi-view images, thereby explicitly embedding 3D information into a 2D diffusion model. Moreover, works on image-based generation~\cite{realfusion, make-it-3d, magic123} and text-based editing~\cite{instruct-nerf2nerf, vox-e, dreameditor, focaldreamer} are also boosted by utilizing these capabilities. Our first static stage performs text-to-3D generation with our Jacobian-based representation.

\begin{figure*}
    \centering
    \includegraphics[width=\textwidth]{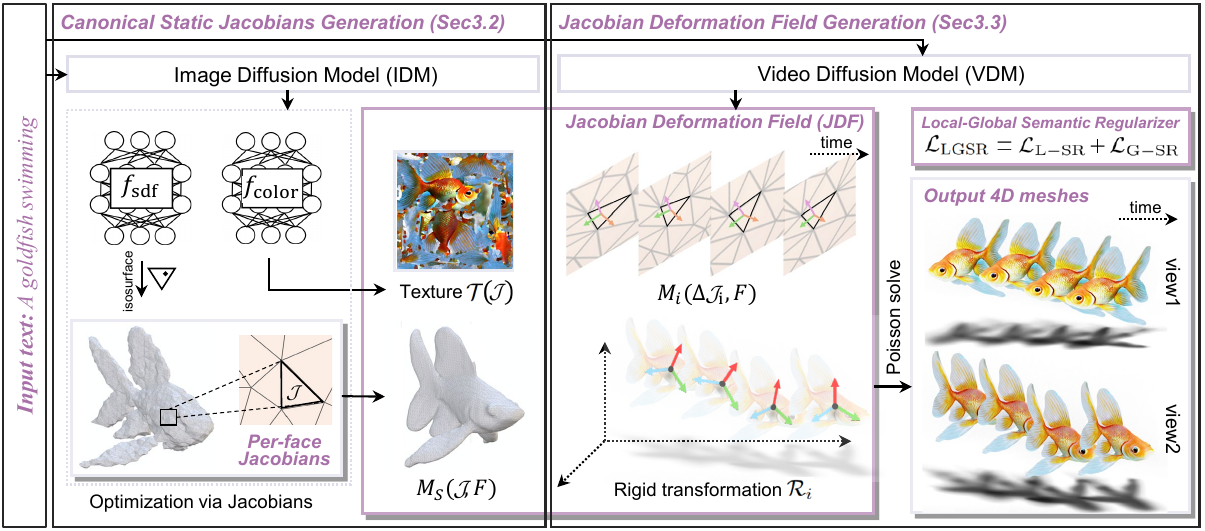}
\caption{
Overview of TextMesh4D. Given a text prompt, we generate a canonical 3D mesh with image diffusion priors and (i) parameterize its dynamics with \textbf{Jacobian Deformation Field (JDF)}, enabling the latter topology-aware deformation. (ii)Video diffusion priors steer the \textbf{JDF evolution} over time, filtering spatially noisy signals into coherent temporal deformations. (iii) A \textbf{Local-Global Semantic Regularizer (LGSR)} further enforces local plausibility and global identity consistency, yielding temporally coherent \textbf{4D meshes}.
}
\label{fig:overview}    
\end{figure*}

\paragraph{Text-to-4D Generation.}
The pioneering effort, MAV3D~\cite{mav3d}, combines a T2V diffusion model with dynamic NeRFs and HexPlane~\cite{hexplane} to optimize both scene appearance and motion consistency. Building on this foundation, 4D-fy~\cite{4d-fy} employs a hybrid SDS approach that integrates T2I, 3D-aware T2I, and T2V diffusion models to achieve high-fidelity 4D generation. Align Your Gaussians (AYG)~\cite{AYG} employs dynamic 3D Gaussian splatting to reduce optimization time while enhancing temporal consistency. 
TC4D~\cite{tc4d} introduces trajectory conditioning to maintain coherence between global and local motion. Although these methods have demonstrated effectiveness, they often rely heavily on NeRF and video models, leading to substantial computational costs. 
Comp4D~\cite{comp4d} employs a Large Language Model (LLM) to segment input prompts into distinct entities, generating 4D objects independently and then combining them based on trajectory data provided by the LLM. Our work pioneers text-to-4D generation using a mesh representation.

\section{Method}

Our goal is to generate 4D meshes from text prompts, using distilled priors from pre-trained diffusion models in a zero-shot manner. The input is a given text prompt describing both the desired object and motion. The output is a sequence of textured 3D meshes, formulated as $\mathcal{M} = \{\mathcal{M}_i = (\mathcal{V}_i, \mathcal{F}, \mathcal{T})\}_{i=1}^{L}$, where $\mathcal{V}_i$ denotes the vertices of the $i$-th mesh, which vary across sequence to capture the motion. $\mathcal{F}$ represents the faces, $\mathcal{T}$ indicates the UV texture map, and $L$ is the length of the sequence. 

We first introduce our proposed Jacobian Deformation Field (Sec. 3.1) and then explain how the total 4D parameters are optimized, covering the static stage (Sec. 3.2) and the dynamic stage (Sec. 3.3).

\subsection{Jacobian Deformation Field}
\label{sec:representation}
 
The total 4D mesh parameters consist of decomposed parts: 1) static parameters for a textured 3D mesh, $\mathcal{M}_s = \{\mathcal{V}_s, \mathcal{F}, \mathcal{T}\}$; 2) a sequence of dynamic parameters $\Theta = \{\theta_i\ = \{\Delta{\mathcal{V}_i}, \mathcal{R}_i\}\}_{i=1}^{L}$, comprising the desired motion including both deformation $\Delta{\mathcal{V}_i}$, and rigid transformation $\mathcal{R}_i$ with global translation and rotation, where $L$ is the length of the sequence. Therefore, the output mesh sequence is $\mathcal{M} = \{\mathcal{M}_i = ( \mathcal{R}_i(\mathcal{V}_s + \Delta{\mathcal{V}_i}), \mathcal{F}, \mathcal{T})\}_{i=1}^{L}$. 

To effectively model the deformation $\Delta{\mathcal{V}_i}$, we shift our modeling unit from vertices to faces. Specifically, we propose Jacobian Deformation Field (JDF), which is defined by per-face Jacobian matrices. The JDF is then composed of canonical static Jacobians and dynamic deformations for time-varying motion. We use $\Delta\mathcal{J}_s$ to denote the static Jacobian offsets and $\Delta\mathcal{J}_i$ to denote the dynamic Jacobian offsets at frame $i$. The optimized canonical Jacobians are defined as
\[
\mathcal{J}^s = \mathcal{I} + \Delta\mathcal{J}_s,
\]
and the dynamic Jacobians are defined as
\[
\hat{\mathcal{J}}_i = \mathcal{J}^s + \Delta\mathcal{J}_i.
\]

\paragraph{Per-face Jacobians.} At each triangle $f_j$ of mesh $\mathcal{M}$, the Jacobian $J_j\in\mathbb{R}^{3\times 3}$ is a linear transformation from the triangle’s tangent space to vertex space $\mathcal{V} \in \mathbb{R}^3$. Defining the deformation as vertex displacement $\Delta\mathcal{V}$ via $\Phi$,
a linear operator $\nabla_j$ is yielded to associate each $\Phi$ with corresponding Jacobian matrix $\nabla_j(\Phi)$. The Jacobian $\nabla_j(\Phi)$ restricts $\Phi$ within each triangle $f_j$, inherently providing low-frequency, smooth signals for deformation as vertex positions.

Given a target assignment of Jacobian $J_j$, a deformation map $\Phi^*$ can be solved following the Poisson equation in a least-squares sense:
\[
\Phi^* \;=\; \min_{\Phi} \sum_{f_j \in \mathcal{F}} |f_j| \,\bigl\lVert \nabla_j(\Phi) \;-\; J_j \bigr\rVert_2^2,
\]
where $|f_j|$ is the area of triangle $f_j$. With deformation map $\Phi^*$ embedding the entire mesh, $\Phi$ can be optimized indirectly by optimizing the Jacobian matrices $\mathcal{J} = \{ J_j \}$ for each face. We then leverage a differentiable Poisson solver layer~\cite{njf} for our optimization.

\paragraph{Jacobian Deformation Field.} 

To achieve high-quality generation, rather than basing on direct vertex positions, we build the parametrization upon Jacobians $\mathcal{J} = \{ J_j \}$ at each triangle as the mesh representation, thereby facilitating smooth, continuous, and globally consistent deformations. Thus, as illustrated in Fig.~\ref{fig:overview}, our method consists of two stages: 1) \textit{Canonical Static Jacobians Generation.} First, we optimize for a high-quality static 3D model $\mathcal{M}_s$. Given the initialized mesh $(\mathcal{V}_0,\mathcal{F})$ extracted from the SDF network, we optimize static Jacobian offsets $\Delta\mathcal{J}_s$, define $\mathcal{J}^s=\mathcal{I}+\Delta\mathcal{J}_s$, and recover the canonical vertices as $\mathcal{V}_s=\mathcal{P}(\mathcal{V}_0,\mathcal{F},\mathcal{J}^s)$ through Poisson integration. This yields $\mathcal{M}_s=\{\mathcal{V}_s,\mathcal{F},\mathcal{T}\}$; 2) \textit{Jacobian Deformation Field Optimization.} With the static parameters fixed, dynamic parameters $\Theta = \{\theta_i\ = \{\Delta{\mathcal{V}_i}, \mathcal{R}_i\}\}_{i=1}^{L}$ are also represented by delta Jacobians as $\Theta = \{\theta_i\ = \{\Delta{\mathcal{J}_i}, \mathcal{R}_i\}\}_{i=1}^{L}$, which are then optimized. The dynamic Jacobians used for frame-wise mesh recovery are then given by $\hat{\mathcal{J}}_i=\mathcal{J}^s+\Delta\mathcal{J}_i$.

\subsection{Canonical Static Jacobians Generation}
\label{sec:static}

\myparagraph{Jacobian Initialization.} Recall that at this stage, our objective is to generate a high-quality, textured 3D mesh solely from an input text prompt. We observe that learning large topological changes via direct mesh initialization (e.g., spot) is challenging for arbitrary inputs and often results in unsatisfactory quality. To this end, we adopt NeuS~\cite{neus} for mesh initialization. {NeuS}~\cite{neus} is a volume rendering method that integrates the advantages of signed distance functions (SDF) and Neural Radiance Fields (NeRF)~\cite{nerf}, better for extracting a 3D geometry and obtaining a mesh as the initialization for further high-quality generation. We denote the NeuS $\mathcal{N} = \{f_{sdf}, f_{color}\}$, both $f_{sdf}$ and $f_{color}$ are networks implemented using MLPs, outputting the SDF and color at point $p$, respectively. The NeuS is then optimized with supervision provided by combined diffusion priors under input text conditioning. After optimization, we extract the surface at the zero-level set of SDF as the initial mesh $\mathcal{M}_0$ using the marching cubes algorithm~\cite{mc}. 

\myparagraph{Optimization via Jacobians.} The extracted mesh $\mathcal{M}_0$ consists of a set of vertices $\mathcal{V}_0 \in \mathbb{R}^{N\times 3}$, faces $\mathcal{F} \in \mathbb{R}^{M\times 3}$, which is converted to a differentiable Jacobian-offset representation. Specifically, we optimize the static Jacobian offsets $\Delta\mathcal{J}_s$, with $\mathcal{J}^s=\mathcal{I}+\Delta\mathcal{J}_s$ initialized as identity Jacobians for optimization. The refined canonical vertices $\mathcal{V}_s$ are recovered through Poisson integration.
We inherit the weights of the color network from the initialization phase and continue to refine them. However, unlike the initialization phase, which utilizes random sampling points, the sampling in this phase is focused on regions near the initialized surface. Thus, we denote the color network at this phase as $f_{color}(\mathcal{J})$. This concentrated sampling strategy allows for a more precise refinement of the color generation. 

\begin{figure}[!t]
    \centering
    \includegraphics[width=0.97\linewidth]{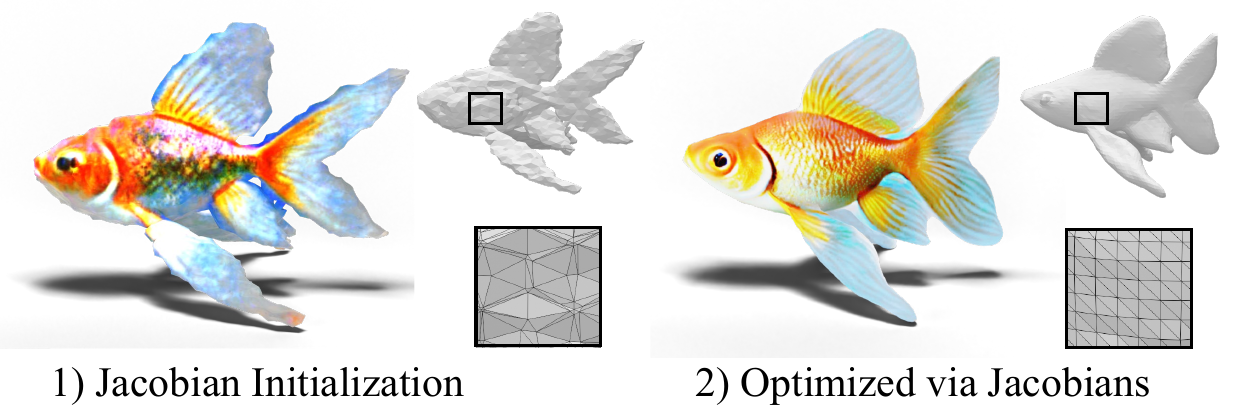}
    \caption{Comparison including geometry and texture between initialization and subsequent jacobian-based generation.}
    \label{fig:coarse_to_fine}
\end{figure}

To achieve high-quality generation, we employ combined diffusion priors from both 3D-aware and 2D-image diffusion models, following~\cite{dream-in-4d, 4d-fy}. 
 The 3D-aware diffusion model, e.g., MVDream~\cite{mvdream}, is trained with multi-view embeddings along with camera parameters, providing a 3D prior and alleviating the Janus problem. As for the 2D-image diffusion priors, we incorporate an additional loss term based on the variational score
distillation (VSD) for appearance improvement.  Combined SDS with them leverages the complementary strengths of 3D-aware and 2D-image diffusion models, resulting in a better generation for our static object:
\begin{equation}
\begin{aligned}
\mathcal{L}_{\mathrm{static}} = \lambda_{3D}\mathcal{L}_{\mathrm{SDS-3D}} + \lambda_{2D}\mathcal{L}_{\mathrm{SDS-2D}},
\end{aligned}
\end{equation}
The loss weights $\{\lambda_{3D}, \lambda_{2D}\}$ are carefully tuned for better generation. They are set to $\{0.7, 0.3\}$ during the initialization stage and adjusted to $\{0.5, 0.5\}$ subsequently. Please refer to Appendix~\ref{subsec:Loss_Details} for loss details. 

To sum up, during the initialization stage, we optimize the networks $\{f_{sdf}, f_{color}\}$. After initialization, with the initialized $\mathcal{M}_0$, parameters that further need to be optimized are $\{\Delta\mathcal{J}_s, f_{color}(\mathcal{J}^s)\}$. Finally, the UV-space texture map $\mathcal{T}(\mathcal{J}^s)$ is extracted from $f_{color}(\mathcal{J}^s)$ for the subsequent motion generation. Fig.~\ref{fig:coarse_to_fine} illustrates the evolution from the initialization phase to the final generation at this stage. 

\subsection{Jacobian Deformation Field Optimization}
\label{sec:dynamic}

With the static parameters fixed, we then optimize the dynamic parameters $\Theta = \{\theta_i\ = \{\Delta{\mathcal{J}_i}, \mathcal{R}_i\}\}_{i=1}^{L}$ to produce the vivid 3D motion. 
Note that a differentiable renderer is required to project the textured mesh sequence with $\mathcal{T}$ to the image space, thus enabling gradient steps during optimization at this stage. We implement the renderer based on NVdiffrast~\cite{laine2020nvdiffrast} as follows:
\begin{equation} \label{eq:rendereq}
    R(\cdot | C) := {\mathcal{M}_i}^{\mathcal{\theta}_i, \mathcal{M}_S} \mapsto I^ {\mathcal{M}_i} \in \mathbb{R}^{H \times W}, %
\end{equation}
where $H$ and $W$ denote the height and width of the rendered image, with $C$ representing the camera extrinsics.

\myparagraph{Objective Function.} The overall objective function at this stage is:
\begin{equation}
\begin{aligned}
\mathcal{L}_{\mathrm{dynamic}} = \mathcal{L}_{\mathrm{VDS}} + \mathcal{L}_{\mathrm{LGSR}}+\mathcal{L}_{\mathrm{others}}
\end{aligned}
\end{equation}

We distill video diffusion priors to provide semantic motion guidance by video score distillation sampling (VDS). This procedure queries a video diffusion model~\cite{zeroscope}, to see how a rendered video from our representation aligns with the input prompt, through the noise sampling of video diffusion process. The gradients are then backpropagated to the dynamic parameters. Please refer to Appendix~\ref{subsec:Loss_Details} for the corresponding loss $\mathcal{L}_{\mathrm{VDS}}$ computation.

However, the stochastic nature of VDS introduces distortions and unstable convergence into the optimization. To address this issue, we design the tailored regularization term $\mathcal{L}_{\mathrm{LGSR}}$, with a synergy of local and global semantic regularity at geometric level, $\mathcal{L}_{\mathrm{L-SR}} + \mathcal{L}_{\mathrm{G-SR}}$, 
for Jacobians' robust optimization under guidance from video score distillation sampling. 

$\mathcal{L}_{\mathrm{G-SR}}$ is a global semantic regularization term on the optimized Jacobians to prevent the divergence too far from the static object's geometry during dynamic optimization, inspired by ~\cite{textdeformer}. This term ensures the global geometry is preserved while still allowing for flexible deformations that capture motion semantics. 

Specifically, the term penalizes the deviation of the dynamic Jacobians $\hat{\mathcal{J}}_i=\mathcal{J}^s+\Delta\mathcal{J}_i$ from the canonical static Jacobians $\mathcal{J}^s$, equivalently regularizing the dynamic Jacobian offsets $\Delta\mathcal{J}_i$:
\begin{equation}
\begin{aligned}
    \mathcal{L}_{\mathrm{G-SR}} 
    = \sum_{i=0}^{\ell-1} \sum_{j=0}^{|\mathcal{F}|-1} 
    e^{\lVert \hat J_{i,j} - J^s_j \rVert_F}
    \lVert \hat J_{i,j} - J^s_j \rVert_F.
\end{aligned}
\label{equ:jaco}
\end{equation}

We then further employ As-Rigid-As-Possible (ARAP) energy~\cite{arap} as the rigidity regularization term $\mathcal{L}_{\mathrm{L-SR}}$:
\begin{equation}
\begin{aligned}
    \mathcal{L}_{\mathrm{L-SR}} = \sum_{i=0}^{\ell-1} \sum_{j=0}^{n-1} \sum_{k \in \mathcal{N}_{v_j}} w_{j, k} || (v_j^i-v_k^i) - R_j(v_j^{s} - v_k^{s}) ||^2, \\
\end{aligned}
\label{equ:arap}
\end{equation}
where $\mathcal{N}_{v_j}$ represents the one-ring neighborhoods of vertex $v_j$. $w_{j,k}=\left( \cot \alpha_{jk} + \cot \beta_{jk} \right)/ 2$, measuring the impact of $v_k$ on $v_j$. $\alpha_{jk}$ and $\beta_{jk}$ are the angles on the faces adjacent to the edge ($v_j$, $v_k$), which are opposite the edge itself. ${R}_j$ represents the optimal rotation estimated by Singular Value Decomposition (SVD)~\cite{arap}. This term encourages the generation to maintain local rigidity during the deformation. 

The Local-Global Semantic Regularizer (LGSR) both preserves local rigidity and maintains global semantic coherence, thereby improving the temporal semantic consistency of the motion, as demonstrated in our ablation studies.
We leave additional details of the other regularizers, e.g., smoothness term and Jacobian's degrees of freedom regularization, in Appendix~\ref{subsec:Loss_Details}.

\section{Experiments}

To evaluate the effectiveness of TextMesh4D, we conduct a comprehensive comparison with state-of-the-art text-to-4D methods across geometric fidelity, appearance quality, motion vividness, semantic consistency, and overall performance. In addition, we benchmark TextMesh4D against representative image-, video-, and 3D-to-4D methods, including both zero-shot and data-driven paradigms, to further assess its capability in 4D content generation. Experimental results demonstrate that TextMesh4D achieves superior performance.

\subsection{Implementation Details}
We summarize our TextMesh4D in Algorithm \ref{alg:ours}. The $\lambda_1$, $\lambda_2$, $\lambda_3$, and $\lambda_4$ in the loss function are respectively set as 0.1, 0.0001, 0.1, and 0.1. We optimize TextMesh4D using Adam~\cite{kingma2015adam} with a learning rate of 0.01, on an NVIDIA RTX A5000 GPU, with an approximate memory consumption of 24GB. The total optimization time for the static stage and the dynamic stage is approximately 1.5 hours.  See Appendix \ref{subsec:Experiment_Details} for additional details.

\begin{algorithm}[t]
\caption{TextMesh4D}
\label{alg:ours}
\small
\textbf{Require:} \\
\hspace*{\algorithmicindent}   $text$ \Comment{input text prompt}\\
\hspace*{\algorithmicindent}   $\mathcal{M} = \{\mathcal{M}_s, \Theta\}$ \Comment{4D representation}\\
\hspace*{\algorithmicindent}   $\mathcal{M}_s = \{\mathcal{V}_s, \mathcal{F}, \mathcal{T}\},\ \mathcal{J}^s=\mathcal{I}+\Delta\mathcal{J}_s$ \Comment{static part}\\
\hspace*{\algorithmicindent}   $\Theta = \{\theta_i = \{\Delta\mathcal{J}_i, \mathcal{R}_i\}\}_{i=1}^{L},\ \hat{\mathcal{J}}_i=\mathcal{J}^s+\Delta\mathcal{J}_i$ \Comment{dynamic part}\\
\hspace*{\algorithmicindent}   $N_{\text{stage-1}}, N_{\text{stage-2}}$ \Comment{iterations for each stage}\\
\hspace*{\algorithmicindent}   $\mathcal{L}_{SDS-2D}, \mathcal{L}_{SDS-3D}, \mathcal{L}_{VDS}$ \Comment{SDS losses}\\
\hspace*{\algorithmicindent}   $\mathcal{L}_{\mathrm{G-SR}}, \mathcal{L}_{\mathrm{L-SR}}, \mathcal{L}_{\mathrm{smooth}}, \mathcal{L}_{\mathrm{dof}}$ \Comment{regularization terms}

\begin{algorithmic}[1]
   \vspace{1em}
    \State // \textbf{Stage 1}
    \State Initialize $\mathcal{M}_0(\mathcal{V}_0, \mathcal{F})$ by NeuS
    \State Parameterize $\mathcal{M}_s$ by static Jacobian offsets $\Delta\mathcal{J}_s$, with $\mathcal{J}^s=\mathcal{I}+\Delta\mathcal{J}_s$ and $\mathcal{V}_s=\mathcal{P}(\mathcal{V}_0,\mathcal{F},\mathcal{J}^s)$
    \For{$\text{iter} \in N_{\text{stage-1}}$}  \Comment{static update}
        \State $\mathbf{grad} = \nabla_{\Delta\mathcal{J}_s, \mathcal{T}}  \mathcal{L}_{SDS-2D}
        + \nabla_{\Delta\mathcal{J}_s, \mathcal{T}} \mathcal{L}_{SDS-3D} +\lambda_0 \nabla_{\Delta\mathcal{J}_s, \mathcal{T}} \mathcal{L}_{\mathrm{dof}}$
        \State UPDATE ($\mathbf{grad}$)
    \EndFor

    \vspace{1em}
    \State // \textbf{Stage 2}
    \State Parameterize dynamic Jacobians as $\hat{\mathcal{J}}_i=\mathcal{J}^s+\Delta\mathcal{J}_i$, with $\Theta = \{\theta_i = \{\Delta\mathcal{J}_i, \mathcal{R}_i\}\}_{i=1}^{L}$
    \For{$\text{iter} \in N_{\text{stage-2}}$} \Comment{dynamic update}
        \State $\mathbf{grad} = \nabla_{\Theta} \mathcal{L}_{VDS} + \lambda_1 \nabla_{\Theta} \mathcal{L}_{\mathrm{G-SR}} + \lambda_2 \nabla_{\Theta} \mathcal{L}_{\mathrm{L-SR}} + \lambda_3 \nabla_{\Theta} \mathcal{L}_{\mathrm{smooth}} + \lambda_4 \nabla_{\Theta} \mathcal{L}_{\mathrm{dof}}$
        \State UPDATE ($\mathbf{grad}$)
    \EndFor
\end{algorithmic}

\end{algorithm}

\subsection{Experimental Setup}

\paragraph{Evaluation Baselines.} 
Evaluation baselines fall into two categories. (1) \textbf{Zero-shot methods}, where we primarily compare against state-of-the-art approaches that distill diffusion priors: (i) NeRF-based methods, including 4D-fy~\cite{4d-fy}, Dream-in-4D~\cite{dream-in-4d}, and TC4D~\cite{tc4d}; and (ii) 3DGS-based methods, including AYG~\cite{AYG} and DG4D~\cite{dreamgaussian4d}. (2) \textbf{Data-driven methods}, included to provide broader comparisons, including Puppeteer~\cite{song2025puppeteer} (3D-to-4D), STAG4D~\cite{STAG4D} (text/video-to-4D), 3-to-4D~\cite{rahamim2024bringing}, and L4GM~\cite{ren2024l4gm} (video-to-4D).

\begin{table*}[!t]
\centering
  \renewcommand{\arraystretch}{1.}
  \renewcommand{\tabcolsep}{3.pt}
  \caption{Quantitative comparisons with baselines. The methods are evaluated in terms of \textbf{CLIP Score, GPT-4V selection}, and metrics of the \textbf{user study.} Note that we also include \textbf{non-text-input} and \textbf{non-zero-shot} baselines for broader comparisons.}
\scalebox{0.81}{
\label{table:4D_results}
\begin{tabular}{lccc|ccccc|ccccc|c}
\toprule
 &   &   &   &  \multicolumn{5}{c}{\textbf{\textit{GPT-4V Selection (\%)}}}   &   \multicolumn{5}{c}{\textbf{\textit{User Studies}}}  & \\
 \textbf{Method} & \textbf{Geometry} & \textbf{Condition} & \textbf{CLIP}$\uparrow$ & AQ$\uparrow$ & SQ$\uparrow$ & MQ$\uparrow$ & TA$\uparrow$ & Overall$\uparrow$ & AQ$\uparrow$ & SQ$\uparrow$ & MQ$\uparrow$ & TA$\uparrow$ & Overall$\uparrow$  & \textbf{GPU requirement} \\
\midrule
\textbf{4D-fy} & \cellcolor{gray}NeRF & \cellcolor{gray}Text &31.03 & 5.68 & 8.02 & 3.57 & 8.42 & 5.36  & 3.1 & 2.9 & 1.5 & 2.6 & 2.4 & 1 * A100 80GB \\
\textbf{Dream-in-4D} & \cellcolor{gray}NeRF & \cellcolor{gray}Text/Image &31.67 & 4.22 & 7.71 & 6.19 & 10.36 & 6.03  & 2.3 & 2.9 & 2.6 & 3.2 & 2.7  & 1 * A100 80GB \\
\textbf{TC4D} & \cellcolor{gray}NeRF & \cellcolor{gray}Text &31.83 & 4.95 & 7.40 & 3.81 & 6.79 & 5.58  & 2.7 & 2.9 & 1.6 & 2.1 & 2.5 & 1 * A100 80GB  \\
\textbf{AYG} & \cellcolor{gray}Gaussian primitive& \cellcolor{gray}Text &30.14 & 6.23 & 8.24 & 9.05 & 11.65 & 8.04  & 3.4 & 3.1 & 3.8 & 3.6 & 3.6  & 128 * A100 80GB \\
\textbf{DG4D} & \cellcolor{gray}Gaussian primitive & \cellcolor{gray}Text &28.70 & 2.20 & 2.92 & 3.33 & 5.18 & 2.90  & 1.2 & 1.1 & 1.4 & 1.6 & 1.3  &  1 * A100 80GB \\
\midrule
\textbf{STAG4D} & \cellcolor{gray}Gaussian primitive & \cellcolor{gray}Text/Video &31.93 & 5.13 & 6.91 & 4.53 & 11.33 & 6.92  & 2.8 & 2.6 & 1.9 & 3.5 & 3.1  & 1 * 3090 24GB \\
\textbf{L4GM} & \cellcolor{gray}Gaussian primitive & \cellcolor{gray}Video & N/A & 4.48 & 9.11 & 8.29 & N/A & 7.69  & 2.9 & 3.4 & 2.7 & N/A & 3.2 & 128 * A100 80GB \\
\textbf{3-to-4D} & \cellcolor{gray}NeRF & \cellcolor{gray}Text+3D & 29.62 & 4.22 & 7.44 & 7.39 & 7.81 & 6.03  & 2.3 & 2.8 & 3.1 & 2.4 & 2.7  &  1 * A100 80GB\\
\textbf{Puppeteer} & \cellcolor{gray}Mesh & \cellcolor{gray}3D & N/A & 6.05 & 10.10 & 8.10 & N/A & 8.26  & 3.3 & 3.8 & 3.4 & N/A & 3.7  & 8 * A100 80GB \\
\midrule
\textbf{Ours} & \cellcolor{gray}Mesh & \cellcolor{gray}Text &\textbf{32.32} & \textbf{56.82} & \textbf{32.16} & \textbf{45.70} & \textbf{38.45} & \textbf{43.19} & \textbf{4.8} & \textbf{4.2} & \textbf{4.6} & \textbf{4.3} & \textbf{4.5} & \textbf{1 * A5000 24GB}\\
\bottomrule
\end{tabular}
}

\end{table*}

\begin{figure*}[t]
    \centering
    \begin{minipage}{0.58\linewidth}
        \centering
        \includegraphics[width=0.99\linewidth]{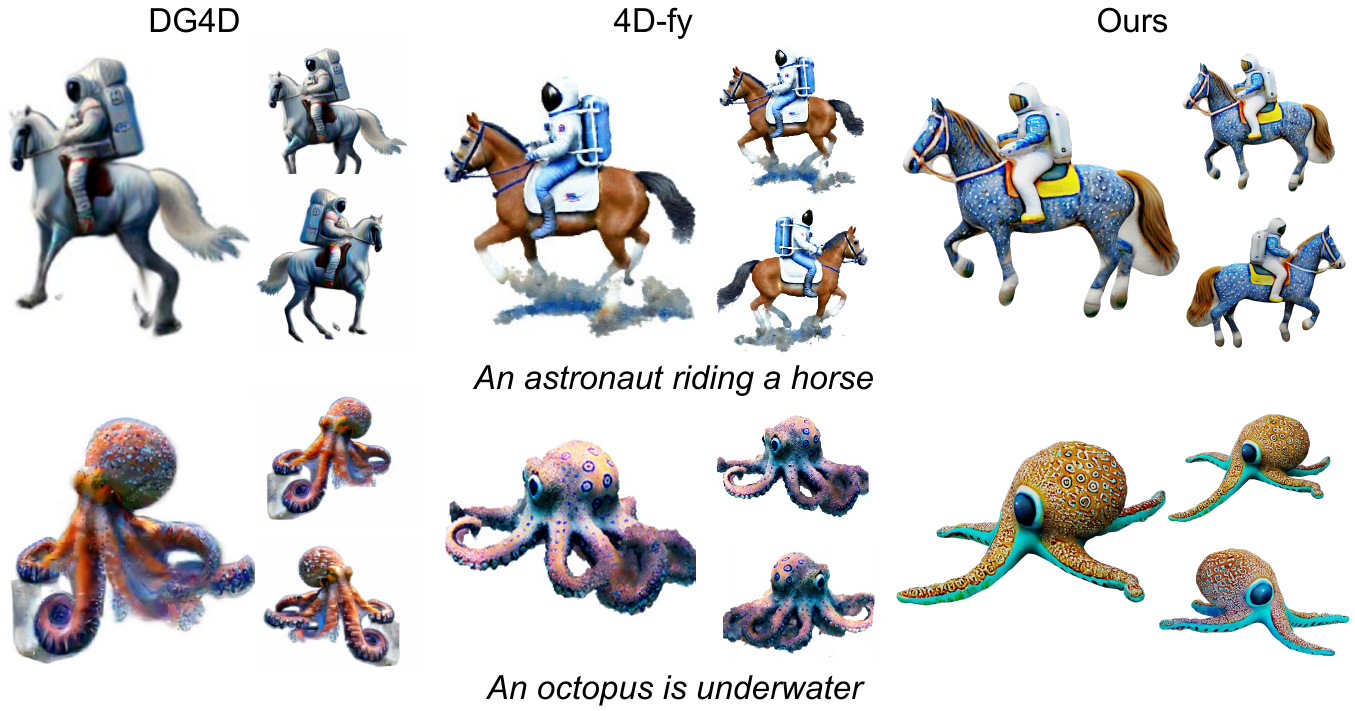}
        \caption{Overall 4D comparison results.}
        \label{fig:compare-overall}
    \end{minipage}%
    \hspace{2mm}
    \begin{minipage}{0.4\linewidth}
        \centering
        \includegraphics[width=0.95\linewidth]{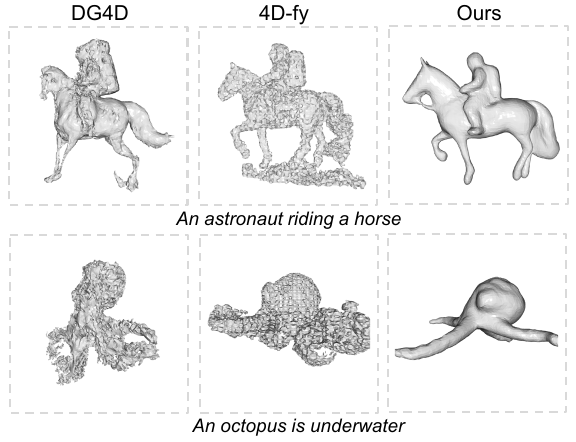}
        \caption{Underlying geometry comparison.}
        \label{fig:compare-geometry}
    \end{minipage}
\end{figure*}

The evaluation prompts are the union of prompt sets from the baseline papers. All baseline results were generated using their official GitHub repositories and default configurations. For methods with unavailable code (e.g., AYG), we use videos from their official project page for qualitative comparison. Note that we exclude CT4D~\cite{chen2024ct4d} from our quantitative comparisons, as its official implementation and results were not publicly available at the time of our experiments. However, we provide a qualitative visual comparison in Appendix \ref{subsec:Com_CT4D} by using images directly from the paper.

\paragraph{Evaluation metrics.} Quantitatively evaluating text-to-4D generation remains challenging due to the lack of ground-truth data. To enable a comprehensive and reliable quantitative evaluation, we adopt evaluation metrics comprising CLIP score, GPT-4V selection, and perceptual user studies, collectively covering multiple quality dimensions.

\subsection{Qualitative Evaluations}
\myparagraph{Diverse results.} In Fig.~\ref{fig:diverse} and Fig.~\ref{fig:more_res} (in Appendix), we present a diverse collection of our generated 4D results, which demonstrates our method's capability in generating diverse and complex motions. Our TextMesh4D successfully handled various rigid or non-rigid deformations, including articulated animal motions (butterfly) or human motions (knight), robotic motions (robotic arm), soft-body deformations (snake), and fluid-like behaviors (flame).

\paragraph{Comparison results.}

Our qualitative comparison focuses on the underlying geometry representation of 4D content, as this is central to our method's contribution. To reveal the geometric fidelity of different 4D representations, we select representative baselines with publicly available code, making their underlying geometry accessible. We adopt 4D-fy~\cite{4d-fy}, a publicly available NeRF-based method for text-to-4D generation, and DreamGaussian4D~\cite{dreamgaussian4d}, a 3DGS-based method for image-to-4D generation. For fairness, we input text prompts into a SOTA text-to-image model to generate high-quality images, which then serve as inputs for DreamGaussian4D. Together with our mesh-based approach, this selection ensures our qualitative comparison covers the mainstream geometry representations: NeRF, 3DGS, and mesh.

\begin{figure*}[t]
    \centering
    \includegraphics[width=0.93\linewidth]{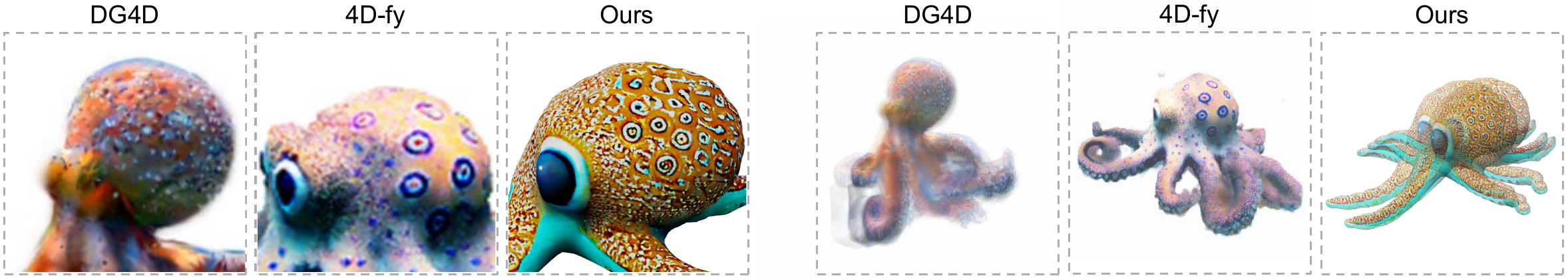}
    \caption{Zoom-in details (left) and motion-centric comparison (right), where opacity conveys motion amplitude: lower opacity indicates larger motion, and vice versa.}
    \label{fig:compare-detail}
\end{figure*}

As shown in Figs.~\ref{fig:compare-overall} (overall comparison), \ref{fig:compare-geometry} (geometric preservation), and \ref{fig:compare-detail} (appearance and motion vividness), we provide a comprehensive comparison of 4D generation results. Our method significantly outperforms the baselines across all evaluation dimensions. The implicit representations used by the baselines, specifically NeRF in 4D-fy and 3D Gaussian Splatting (3DGS) in DreamGaussian4D, cause most of the generated motion to manifest as floating artifacts in the geometric space, making the results visually ambiguous and difficult to interpret. Moreover, because these methods do not model global transformations, the generated motion remains confined to local object regions, which further limits their expressive capability. In contrast, our method employs a mesh representation that integrates both local deformations and global transformations, enabling effective spatial movement solely conditioned on input text.

\subsection{Quantitative Evaluations}

\myparagraph{CLIP Score.} Following common practice, we compute the CLIP score~\cite{CLIP} to measure the similarity between input text prompts and the corresponding generated results, where a higher score indicates better alignment with input descriptions. Multiple camera views and frames over time are sampled during the computation of the CLIP score. As shown in Table~\ref{table:4D_results}, TextMesh4D achieves the highest mean CLIP score across evaluation prompts among all baselines.

\myparagraph{User study.} We conduct perceptual user studies to evaluate sample quality along the dimensions of \emph{appearance quality (AQ)}, \emph{3D structure quality (SQ)}, \emph{motion quality (MQ)}, \emph{text alignment (TA)}, and \emph{overall preference (Overall)}. A total of 31 participants were recruited for evaluation. Details are provided in Appendix \ref{subsubsec:Experiment_Details}, and the numerical results are reported in Table ~\ref{table:4D_results}, which consistently favor our method over the baselines in direct comparisons.

\myparagraph{GPT-4V selection.} 
Moreover, we follow InterFusion~\cite{interfusion} and further leverage the advanced image understanding capabilities of GPT-4V to enable a more fine-grained evaluation. Specifically, we prompt GPT-4V to select the preferred result based on the same criteria as used in the aforementioned user study. No in-context examples are provided during prompting to maintain a zero-shot setting. Results are reported in Table~\ref{table:4D_results}. GPT-4V provides a complementary evaluation to human feedback and further validates the superiority of our method.

\begin{table}[!t]
    \caption{Quantitative ablation results by GPT-4V.}
    \setlength{\abovecaptionskip}{10pt}
    \renewcommand{\arraystretch}{0.95} 
    \label{tab:ablation}
    \begin{center}
    \resizebox{0.95\columnwidth}{!}{
    \begin{tabular}{l|cccc|c}
        \toprule
         &  & \multicolumn{4}{c}{\textbf{\textit{GPT-4V Selection (\%)}}}\\
        \textbf{Settings} & AQ & SQ & MQ & TA & Overall \\\midrule
        Vertex (Var-A)  & 1.9 & 1.0 & 1.5 & 1.8 & 1.1 \\
        Vertex (Var-B) & 1.1 & 7.3 & 0.5 & 1.3 & 3.9 \\
        w/o G-SR  & 15.2 & 9.0 & 3.4 & 5.6 & 7.4 \\
        w/o L-SR  & 5.6 & 6.9 & 39.7 & 26.3 & 10.3 \\
        \textbf{Ours (JDF+LGSR)}  & \textbf{76.2} & \textbf{75.8} & \textbf{55.0} & \textbf{66.0} & \textbf{77.3}\\
        \bottomrule
    \end{tabular}}
    \end{center}
\end{table}

\begin{figure}[!t]
    \centering
    \includegraphics[width=\linewidth]{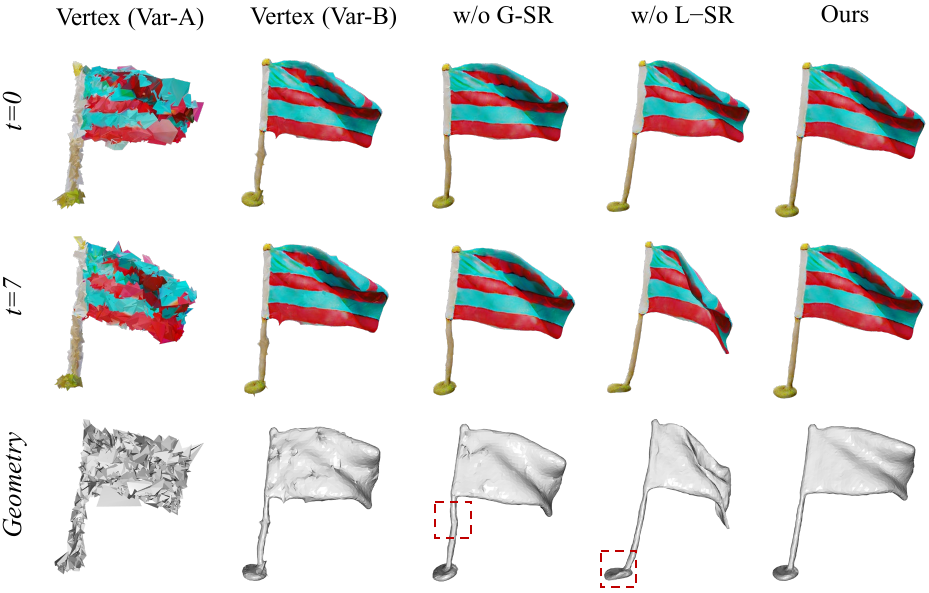}
    \caption{Qualitative ablations with the given text prompt ``a flag fluttering in the air''. The top two rows showcase two frames, while the bottom row illustrates the geometry corresponding to the second row.}
    \label{fig:ablation}
    \vspace{-5pt}
\end{figure}

\subsection{Ablation Study}

\myparagraph{Effectiveness of Jacobian Deformation Field (JDF).} We conduct ablation studies to validate the critical role of our proposed JDF in addressing deformation inflexibility. We compare our full method (Ours) against two ablated variants: 1) Vertex (Var-A), which directly optimizes vertex displacements as in prior works, and 2) Vertex (Var-B), which applies an intermediate rigging structure to vertex displacements for deformation enhancement. 

As illustrated in Fig.~\ref{fig:ablation}, ablation results (1st and 2nd columns) clearly demonstrate the limitations of vertex-based optimization. The Vertex (Var-A) variant suffers from severe geometric shattering and topological errors, as direct vertex optimization is overly constrained by mesh connectivity, making it incapable of expressing complex deformations. The Vertex (Var-B) variant mitigates the surface-breaking artifacts. However, it exhibits overly rigid dynamics (the unnatural flagpole), and geometric artifacts are still present. In contrast, our JDF method generates a smooth, continuous, and visually plausible deformation. This experiment validates our core insight: by shifting the deformation unit from vertices to faces and modeling transformations via JDF, we effectively release the deformation from rigid topological constraints. This allows our method to achieve both high-fidelity geometric integrity and the expressive flexibility required for complex motions.

\myparagraph{Effectiveness of Local-Global Semantic Regularizer (LGSR)} We also conduct ablation studies to validate our Local-Global Semantic Regularizer (LGSR) in addressing semantic inconsistency, which arises from noisy priors from text-to-video (T2V) models that cause both local surface corruption and global identity drifts. We compare our full method against two ablated variants: 1) w/o G-SR, which applies only the local regularizer, and 2) w/o L-SR, which applies only the global regularizer.

As shown in Fig.~\ref{fig:ablation}, ablation results (3rd and 4th columns) demonstrate that both local and global components are indispensable. The w/o L-SR variant highlights the necessity of the local component (L-SR), which provides local rigidity to stabilize local geometry against noisy priors. Without it, the geometry appears stretched like clay during motion (highlighted in red). Conversely, the w/o G-SR variant showcases that the global component (G-SR) is essential for preserving global semantic identity. Without this component, the model fails to preserve the flagpole's identity (highlighted in red). Our full JDF+LGSR (Ours), leveraging both regularizers, works in tandem to suppress both local and global noise from the T2V prior at the geometric level. This strikes a balance between rigidity and flexibility, and generates natural and high-fidelity results.

\section{Discussion}

\myparagraph{Failure Cases and Boundary Analysis.}
TextMesh4D assumes fixed mesh connectivity $\mathcal{F}$ throughout the generated sequence, and is therefore mainly suitable for continuous fixed-topology motion, including articulated motion, soft-body deformation, cloth-like dynamics, and fluid-like visual dynamics under fixed connectivity. However, it may fail on prompts that require explicit topology changes, such as object splitting, merging, breaking, or the emergence of new parts. Such topology-changing prompts, e.g., shattering, exploding, or growing new parts, cannot be faithfully represented by our fixed-topology formulation. In these cases, TextMesh4D tends to produce over-smoothed deformation, surface stretching, or incomplete semantic motion.

\myparagraph{Limitations \& Future Work.}
As our framework follows a two-stage process, where static content is generated first, followed by dynamic motion synthesis, the quality of dynamic generation depends on the success of the static stage. If the static composition is unsatisfactory, errors may propagate into subsequent motion synthesis and lead to degraded 4D results. We believe this issue could be alleviated with further advances in 3D generation techniques.

Furthermore, the optimization space of global transformations is constrained by the camera space of differentiable rendering. Combining our framework with the latest video diffusion models under explicit camera control offers a promising direction. Lastly, due to the inherent trade-off between motion stability and diversity, our current design prioritizes generalizable and stable motion, which limits its ability to handle highly exaggerated or topology-changing dynamics as discussed above. Addressing such cases will be an interesting avenue for future exploration.

\section{Conclusion}

We present TextMesh4D, a novel framework for generating dynamic 3D mesh sequences from text. By shifting the deformation unit from vertices to faces via the Jacobian Deformation Field (JDF), our method overcomes the inherent deformation inflexibility of mesh topology. Furthermore, our proposed Local-Global Semantic Regularizer (LGSR) effectively suppresses stochastic semantic inconsistency introduced by video diffusion priors, ensuring both local motion coherence and global identity consistency. Extensive experiments validate that TextMesh4D sets a new state of the art in temporal stability, geometric fidelity, and visual quality for text-to-4D generation, demonstrating the viability of mesh-centric approaches in 4D content creation.

\section*{Acknowledgements}

We thank anonymous reviewers for their constructive feedback. This work was supported in part by the NSFC (62325211, 62132021), the Research Fund of Jiangsu Key Laboratory of AI for Industries (E6420016G8).

\section*{Impact Statement}
This paper presents work whose goal is to advance the field of machine learning. There are many potential societal consequences of our work, none of which we feel must be specifically highlighted here.

\clearpage

\bibliography{icml2026}
\bibliographystyle{icml2026}

\clearpage
\onecolumn
\appendix
\section*{Appendix}

\begin{figure*}[!b]
    \centering
    \includegraphics[width=\linewidth]{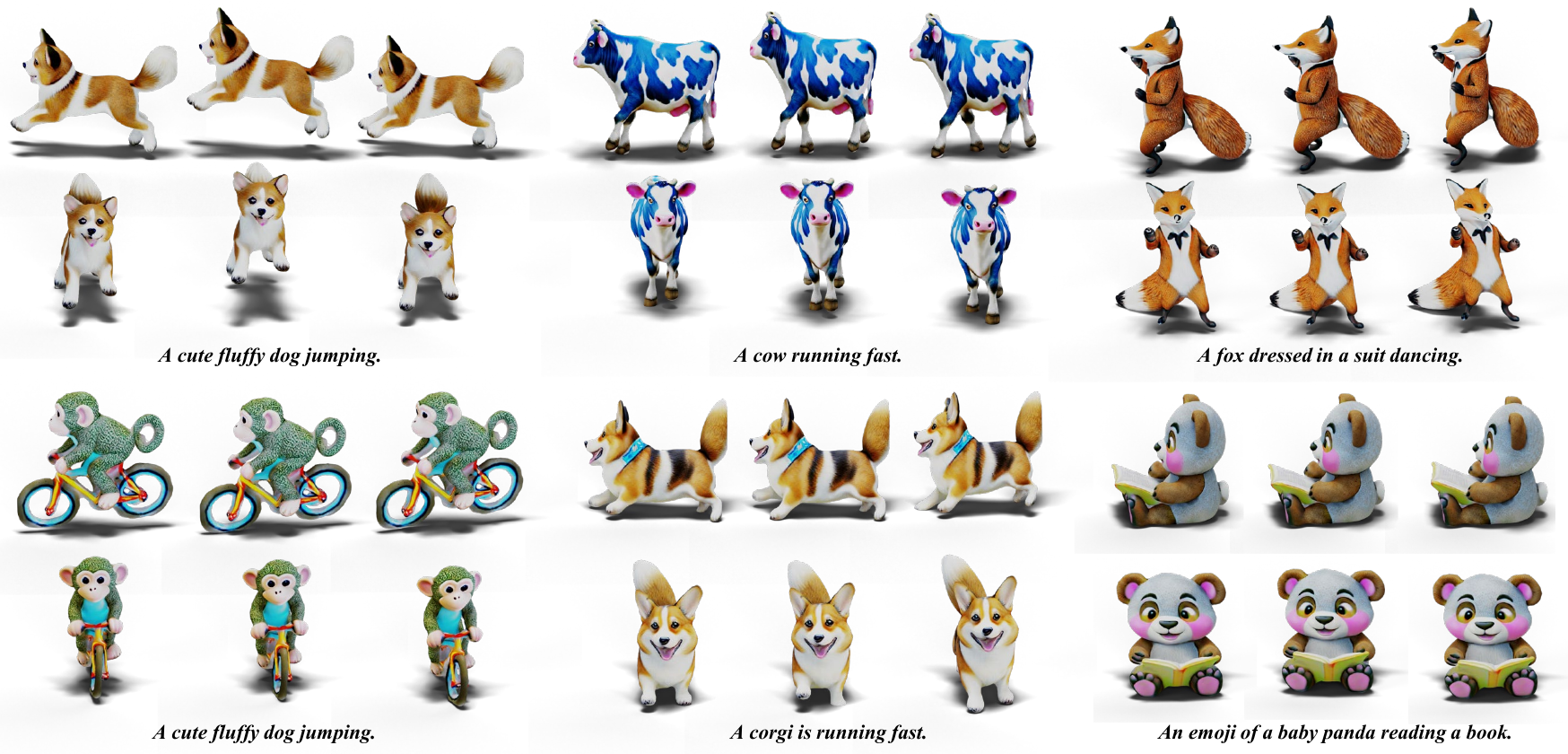}
    \caption{Additional results of TextMesh4D, showcasing consistency and realistic motion. Shown here are intermediate frames from the generated motion sequences.}
    \label{fig:more_res}
\end{figure*}

\subsection *{A. Experiment Details}
\makeatletter
\def\@currentlabel{A}
\makeatother
\label{subsec:Experiment_Details}

\subsubsection *{A1. Baseline Implementation}

We benchmark our method against 9 baselines: 4D-fy~\cite{4d-fy}, Dream-in-4D~\cite{dream-in-4d}, TC4D~\cite{tc4d}, AYG~\cite{AYG}, DG4D~\cite{dreamgaussian4d}, STAG4D~\cite{STAG4D}, L4GM~\cite{ren2024l4gm}, 3-to-4D~\cite{rahamim2024bringing}, and Puppeteer~\cite{song2025puppeteer}. We utilize the official implementations for most baselines. Notably, for AYG, we compare directly against the results shown in their demo videos due to code unavailability. For 3-to-4D, we faithfully re-implemented the method following the original paper, as their released code is incomplete.

\subsubsection *{A2. User Study Details}
\makeatletter
\def\@currentlabel{A2} 
\makeatother
\label{subsubsec:Experiment_Details}
We conduct the user study to ensure a convincing comparison. Evaluators are asked to compare generated videos with the corresponding text prompt, and provide their preferences across the following dimensions: \emph{Appearance Quality (AQ)}, \emph{Structure Quality (SQ)}, \emph{Motion Quality (MQ)}, \emph{Temporal Alignment (TA)}, and \emph{Overall}. The following instructions of the dimensions are given to the evaluators:

\begin{itemize}
    \item \textbf{Appearance Quality:} Evaluate the clarity and visual appeal as it appears from any particular viewpoint (ignoring, e.g., inconsistencies in appearance across different viewpoints). Your assessment should focus on the appearance of the foreground object and ignore the background of the video.
    \item \textbf{Structure Quality:} Assess the details and realism of the shape across the multiple viewpoints shown in the video. 
    \item \textbf{Motion Quality:} Assess the realism of motion, including the amount of motion and how naturally the movements in the video are portrayed. 
    \item \textbf{Text Alignment:} Determine how accurately each video reflects the content of the text prompt. Consider whether the key elements of the prompt are represented.
    \item \textbf{Overall Preference:} State your overall preference between the two videos. This is your subjective appraisal of which video, in your view, stands out as better based on appearance quality, 3D structure quality, motion quality, and text alignment, (i.e., overall quality).
\end{itemize}

We received responses from 31 participants (mean age = 26.3 ± 1.5 years) across 48 prompts, rating each on a scale of 1 to 5 across above 5 dimensions. For our comparison against AYG, we adopted all the prompts from their open-source website. Example prompts used in the survey are as follows:

\begin{center}
\begin{minipage}{0.9\linewidth}
{\footnotesize
\noindent\rule{\linewidth}{0.8pt}

{\itshape
\begin{multicols}{2}
\begin{enumerate}
    \setlength{\itemsep}{0pt}
    \setlength{\parsep}{0pt}
    \setlength{\topsep}{0pt}

    \item A bee fluttering its wings fast.
    \item A cat singing, best quality.
    \item A corgi is running fast.
    \item A cow running fast.
    \item A dalmatian is running fast.
    \item A dog riding a skateboard.
    \item A fox dressed in a suit dancing.
    \item A llama running fast.
    \item A space shuttle launching.
    \item A squirrel riding a motorcycle.
    \item A storm trooper walking forward and vacuuming, best quality.
    \item A tiger running fast.
    \item An ancient roman statue dancing, full body, portrait, game, unreal.
    \item An octopus is under water.
    \item Beer pouring into a glass.
    \item Chihuahua running.
    \item Dragon armor fluttering its wings fast.
    \item Flying dragon on fire.
    \item Mage in purple robe dancing, full body, portrait, game, unreal.
    \item Mage in purple robe running forward, full body, portrait, game, unreal.
    \item Santa Claus carrying a big bag is walking forward, portrait, game, unreal.
    \item Sorcerer in blue robe dancing, full body, portrait, game, unreal.
    \item Sorcerer in blue robe running forward, full body, portrait, game, unreal.
    \item Wood on fire.
    \item An astronaut is playing the electric guitar, ultra realistic.
    \item Assassin with sword running fast, portrait, game, unreal.
    \item Beautiful, intricate butterfly flutters its wings.
    \item Flamethrower on fire, scifi, cyberpunk, photorealistic.
    \item Tesla trooper shooting lightning, body keep still, scifi, game, character, photorealistic.
    \item An astronaut riding motorcycle.
    \item A knight in shining armor holding a sword and shield fighting.
    \item A monkey is playing bass guitar, best quality.
    \item A panda dancing.
    \item An emoji of baby panda reading a book.
    \item A panda surfing a wave, best quality.
    \item Poseidon holding his trident emerging from the sea.
    \item A humanoid robot playing the violin.
    \item An astronaut riding a horse, best quality.
    \item A dog wearing a Superhero outfit with red cape flying through the sky.
    \item A dog swimming in ocean, ocean waves crashing against the dog.
    \item A donkey running fast.
    \item Clown fish swimming.
    \item Waves crashing against a lighthouse.
    \item A squirrel playing on a swing set, best quality.
    \item Tesla trooper standing upright from half squat position, scifi, game, character, photorealistic.
    \item A turtle swimming.
    \item A purple unicorn flying.
    \item Volcano eruption.
\end{enumerate}
\end{multicols}
}

\noindent\rule{\linewidth}{0.8pt}
}
\end{minipage}
\end{center}

\begin{figure}[!t]
    \centering
    \makebox[\linewidth][l]{%
        \hspace*{0.248\linewidth}%
        \includegraphics[width=0.475\linewidth]{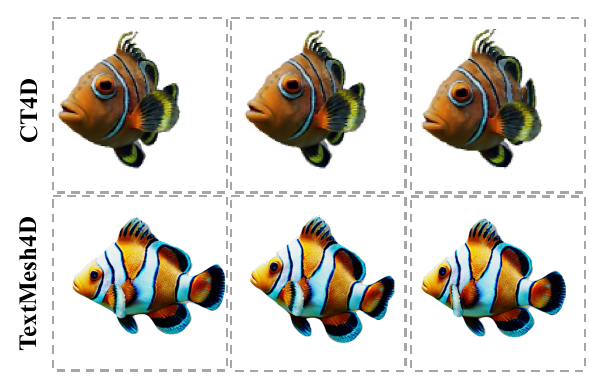}
    }
    \caption{Qualitative comparison with CT4D, under the text prompt ``a clown fish swimming''.}
    \label{fig:ct4d}
\end{figure}

\subsection *{B. Comparison with CT4D}
\makeatletter
\def\@currentlabel{B}
\makeatother
\label{subsec:Com_CT4D}
Due to the unavailability of the CT4D code, we used the image results provided in their paper for comparison. We employed GPT-4V for quantitative evaluation, and the results are summarized in Table~\ref{tab:result}. Additionally, we provide a qualitative visual comparison in Figure ~\ref{fig:ct4d}.

\subsection *{C. Loss Function Details}
\makeatletter
\def\@currentlabel{C}
\makeatother
\label{subsec:Loss_Details}

\subsubsection *{C1. SDS (Score Distillation Sampling)}
SDS (Score Distillation Sampling) has been introduced by DreamFusion \cite{dreamfusion}. By injecting the sampled noise $\epsilon$ to $\renderedimage$ at time step t, the noisy image $\renderedimage_{\timestep}$ is produced. The pre-trained 2D text-to-image diffusion model ${\phi}$ provides a denoising network $\hat{\epsilon}_{\phi}(\renderedimage_{\timestep} ; y, t)$ that predicts the noise $\hat{\epsilon}$ given the noisy image $\renderedimage_{\timestep}$, time step t, and text embedding y. SDS then optimizes the scene parameters ${\psi}$ by minimizing the difference between the predicted noise and the added noise:
\begin{equation}
\nabla_{\modelparams} \mathcal{L}_{\mathrm{SDS-2D}} = \mathbb{E}_{\timestep, \noise} [w(\timestep) (\hat{\noise}_{\noisepredictnet}(\renderedimage_{\timestep}; \timestep, \textembedding)-\noise) \frac{\partial \renderedimage}{\partial \modelparams}], 
\label{equ:SDS-2D}
\end{equation}
Here, \( \timestep \) represents the timestep in the reverse diffusion process, as well as the noise level, $w(t)$ is the weighting term at $t$, \( y \) is the text prompt, and \( c \) represents the camera extrinsics.

Classical SDS gradient is derived from single-view 2D pre-trained diffusion models. By replacing the single-view model with a 3D-aware model~\cite{mvdream} that incorporates multi-view information along with camera parameters during training, we obtain the 3D-aware SDS:
\begin{equation}
\nabla_{\modelparams} \mathcal{L}_{\mathrm{SDS-3D}} = \mathbb{E}_{\timestep, \noise, \cameraextrinsics} [w(\timestep) (\hat{\noise}_{\noisepredictnet}(\renderedimage_{\timestep}; \timestep, \textembedding, \cameraextrinsics)-\noise) \frac{\partial \renderedimage}{\partial \modelparams}].
\label{equ:SDS-3D}
\end{equation}
Here, \( \timestep \) represents the timestep in the reverse diffusion process, \( w \) is the weight function, \( \hat{\noise} \) denotes the predicted noise from the model, 

\subsubsection *{C2. VSD (Variational Score Distillation)} Variational Score Distillation (VSD), as introduced in \cite{prolificdreamer}, is designed to enhance the visual quality of images rendered from a scene. It achieves this by leveraging a pre-trained text-to-image model along with a fine-tuning strategy that pushes image quality beyond the capabilities of a standalone 3D-aware text-to-image model. We augment the standard SDS gradient with the output of an additional text-to-image diffusion model, which is fine-tuned using low-rank adaptation
during optimization. The VSD gradient is given by:
\begin{equation}
\begin{aligned}
\nabla_{\modelparams} \mathcal{L}_{\mathrm{VSD}} =\;& \mathbb{E}_{t,\boldsymbol{\epsilon},c}[w(t)\Bigl(\hat{\epsilon}_\phi\bigl(\mathbf{z}_t, v, t\bigr)-\hat{\epsilon}^{\prime}_\phi\bigl(\mathbf{z}_t, v, t,c\bigr)\Bigr)\frac{\partial \renderedimage}{\partial \modelparams}].
\end{aligned}
\end{equation}

\begin{table}[!t]
    \label{tab:compare}

    \begin{center}
    \footnotesize
    \begin{tabular}{cccccc}
        \toprule
        \textbf{Metric (\%)} & AQ & SQ & MQ & TA & Overall \\\midrule
        \textbf{Ours}  & \textbf{64.1} & \textbf{79.2} & \textbf{68.4} & \textbf{88.2} & \textbf{80.7} \\
        \bottomrule
    \end{tabular}
    \end{center}
    \caption{\footnotesize Quantitative Comparison with CT4D. The numbers indicate the percentage of GPT-4V preferring our results over CT4D's.}
\label{tab:result}
\end{table}

In this expression, \(\hat{\epsilon}^{\prime}\) represents the noise predicted by the fine-tuned diffusion model that incorporates additional conditioning from the camera extrinsics \(c\). The fine-tuning is performed using the standard diffusion objective:
\begin{equation}
\min_\modelparams\mathbb{E}_{t_d,\boldsymbol{\epsilon},c}\Biggl[\Bigl\|\boldsymbol{\epsilon}_\phi^{\prime}\bigl(\mathbf{z}_{t_d}, v, t_d, c\bigr)-\boldsymbol{\epsilon}\Bigr\|_2^2\Biggr].
\end{equation}

We deviate from the original VSD formulation by not performing simultaneous optimization over multiple scene samples, similar to 4D-fy~\cite{4d-fy}.

\subsubsection *{C3. Other Regularization Terms} 
\myparagraph{Temporal Smoothness Regularization.} To facilitate the temporal smoothness during the generated motion, we introduce the smoothness regularization term to penalize large differences between consecutive frames:
\begin{equation}
\begin{aligned}
    \mathcal{L}_{\mathrm{smooth}} = \frac{1}{(\ell-1)} \sum_{i=0}^{\ell-2} (\frac{\lVert {\hat J}_{i} - {\hat J}_{i+1}\rVert_1}{|f|} + \\
    \lVert {\hat t}_{i} - {\hat t}_{i+1}\rVert_1 + \lVert {\hat r}_{i} - {\hat r}_{i+1}\rVert_1).
\end{aligned}
\label{equ:smooth}
\end{equation}

\myparagraph{Dof Regularization.} During the process of solving the Poisson equation about Jacobians, the global translational degree of freedom remains uniform across all optimized matrices, and any translation during optimization does not affect the computation of each triangle's Jacobian matrix. Therefore, we introduce a constraint term on these degrees of freedom (dof). We enforced this constraint in a straightforward manner by uniformly sampling points on the optimized object and computing the corresponding MSE loss against the initial object:
\begin{equation}
\mathcal{L}_{\mathrm{dof}} = \frac{1}{N}\lVert \mathcal{V}_i - \mathcal{V}_S\rVert_2^2,
\label{eq:loss_dof}
\end{equation}

\end{document}


\title{Supplementary Material for TextMesh4D}

\twocolumn[{%
    \maketitle 
    \begin{center}
        \includegraphics[width=\linewidth]{img/supp_diverse.pdf}
        \captionof{figure}{Additional results of TextMesh4D, showcasing consistency and realistic motion. Shown here are intermediate frames from the generated motion sequences.} 
        \label{fig:more_res}
    \end{center}
}]
\newcommand{\renderer}{g}
\newcommand{\modelparams}{\psi}
\newcommand{\renderedimage}{x}
\newcommand{\camerapose}{\zeta}
\newcommand{\textprompt}{T}
\newcommand{\textembedding}{y}
\newcommand{\timestep}{t}
\newcommand{\noise}{\epsilon}
\newcommand{\noisepredictnet}{\phi} 
\newcommand{\cameraextrinsics}{\mathbf{c}} 
\newcommand{\bodypose}{\theta}
\newcommand{\bodyshape}{\beta}
\newcommand{\spatialpoint}{p}
\newcommand{\allpoints}{P}
\newcommand{\crossentropy}{CE}
\newcommand{\hyperparameter}{\eta}
\newcommand{\distancetoanchor}{d}

\input{sec/X_suppl}


{
    \small
    \bibliographystyle{ieeenat_fullname}
    \bibliography{cvpr2026}
}